# Error-Resilient Machine Learning in Near Threshold Voltage via Classifier Ensemble

**Sai Zhang    Naresh Shanbhag**
Department of Electrical and Computer Engineering
University of Illinois at Urbana-Champaign
Urbana, IL 61801
{szhang12,shanbhag}@illinois.edu

## Abstract

In this paper, we present the design of error-resilient machine learning architectures by employing a distributed machine learning framework referred to as *classifier ensemble* (CE). CE combines several simple classifiers to obtain a strong one. In contrast, centralized machine learning employs a single complex block. We compare the random forest (RF) and the support vector machine (SVM), which are representative techniques from the CE and centralized frameworks, respectively. Employing the dataset from UCI machine learning repository and architectural-level error models in a commercial $45\,\text{nm}$ CMOS process, it is demonstrated that RF-based architectures are significantly more robust than SVM architectures in presence of timing errors due to process variations in near-threshold voltage (NTV) regions ($0.3\,\text{V}$ - $0.7\,\text{V}$). In particular, the RF architecture exhibits a detection accuracy ($P_{det}$) that varies by 3.2% while maintaining a median $P_{det} \geq 0.9$ at a gate level delay variation of 28.9% . In comparison, SVM exhibits a $P_{det}$ that varies by 16.8%. Additionally, we propose an error weighted voting technique that incorporates the timing error statistics of the NTV circuit fabric to further enhance robustness. Simulation results confirm that the error weighted voting achieves a $P_{det}$ that varies by only 1.4%, which is $12\times$ lower compared to SVM.

## 1 Introduction

Emerging applications such as recognition, mining, synthesis (RMS), rely on computationally intensive machine learning algorithms to extract patterns from complex data. Conventionally, these algorithms are deployed on large scale general-purpose computing platforms such as CPU and GPU-based clusters, leading to significant cost in energy. Machine learning algorithms play an important role in enabling in-situ data analytics employing energy-constrained embedded platforms. This stringent energy constraint precludes the use of general-purpose hardware platforms, resulting in much interest in dedicated integrated circuit implementation of machine learning kernels [4, 11]. These implementations have shown to achieve a $252\times$ energy reduction for convolutional neural network-based vision system [4] and a $5.2\times$ throughput enhancement for a k-nearest-neighbor (KNN) engine [11] as compared to implementations on general-purpose platforms.
Conventionally, energy consumption of machine learning implementations is minimized by reducing the computational complexity, precision, and data movement [16]. Such techniques exploit the algorithms' intrinsic robustness to numerical errors. This intrinsic robustness can be exploited in another way to further reduce energy - by implementing such kernels on circuit fabrics that operate at the limits of energy efficiency and hence tend to be unreliable. Examples of such circuit fabrics include near threshold voltage (NTV) CMOS [8], and emerging devices such as CNFET [18] and spin [15]. For example, it has been shown that NTV operation achieves up to $10\times$ energy savings, but leads to increased delay variations as large as $14\times$ [8]. Hardware errors are expected to be common place in scaled or beyond CMOS processes [15, 18], and there is great interest in understanding the behavior of machine learning architectures in presence of these errors.

The most common machine learning architecture is the centralized architecture (see Fig. 1(a)) where a complex block such as the support vector machine (SVM) is employed to process all the input data. However, the computational complexity of centralized architecture increases dramatically as a function of the non-linearity of the decision boundary [17]. The *classifier ensemble* (CE, see Fig. 1(b)) is a distributed architecture for machine learning which combines several weak (low-complexity) classifiers to form a strong classifier. CE enables on-chip training due to its distributed nature, and exhibits robustness to feature/label noise. Thus, it is of great importance to compare the robustness and energy efficiency of distributed machine learning architectures designed using CE with centralized architectures such as SVM.

In this paper, we compare the robustness of distributed and centralized machine learning architectures in presence of timing errors due to NTV operations. Specifically, we compare a CE method - random forest (RF) - with SVM using architectural-level error models [19] in $45\,\text{nm}$ CMOS. Employing the breast cancer data set in the UCI machine learning repository [1], we show that RF achieves a detection accuracy ($P_{det}$) that varies by 3.2% while maintaining a median $P_{det} \geq 0.9$ when operating with a gate level delay variation of 28.9%. This is $5\times$ lower as compared to SVM. We further propose a new error weighted voting to enhance the robustness of RF by employing the timing error statistics of the NTV circuit fabric. Simulation results confirm that the proposed method leads to a $P_{det}$ that varies by only 1.4%, which is $12\times$ lower compared to SVM.

The rest of the paper is organized as follows. Section 2 provides the background for CE, SVM, and the architectural level error models. Section 3 describes architectures for RF and SVM classifiers. Section 4 presents simulation results validating the error models in a $45\,\text{nm}$ CMOS process, and employs these models to compare the detection accuracy of SVM, RF, and proposed RF with error weighted voting scheme. Conclusions are presented in Section 5.

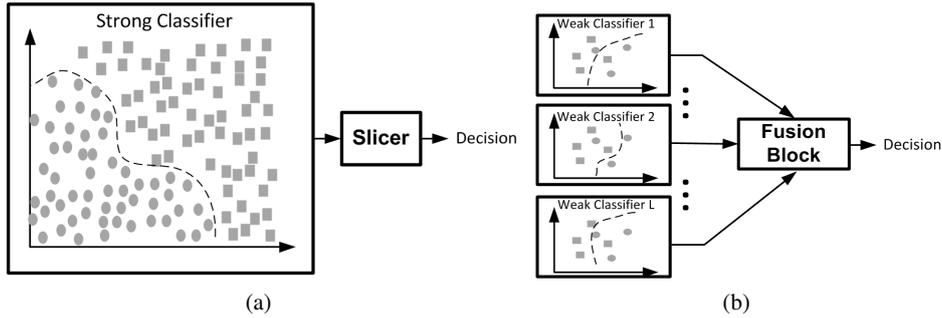

Figure 1: Two distinct machine learning frameworks: a) centralized machine learning, and b) classifier ensemble.

## 2 Background

### 2.1 Classifier Ensemble (CE)

Classifier ensemble (also referred to as Multiple Classifier System) has been employed to enhance the performance of single classifier system [6]. A wide variety of CE methods exist. In bootstrap aggregating (bagging) [2], multiple training sets are generated from the original training set via random sampling with replacement, in order to train multiple classifiers. Adaboost [9] is another popular method for ensemble generation. The training samples are re-weighted after each iteration so that the mis-classified samples get higher weights. Other methods such as randomness injection, random subspace [10] and output coding [6] also exist.

RF is a CE method that combines random subspace and bagging, while employing an ensemble of decision trees (DTs) as weak classifiers. It is a popular technique for classification, prediction, variable selection, and has shown superior results compared to other linear and non-linear predictive modeling techniques [3]. Advantages include parallel training, robustness to overfitting, ease of design, the capability of getting out-of-bag (OOB) error estimate, and others.

In RF, the training set for each individual DT is generated using bagging. During the training of each DT, a random subset of features are selected, and the best feature is selected to split the DT according to an appropriate criterion. Several variations of RF exists based on the type of DT used as base classifiers. Classification and regression tree (CART) [3] employs the Gini index as a measure of the impurity of nodes. ID3 [14] employs information gain as the criterion. C4.5 [14] improves ID3 by using the information gain ratio.

### 2.2 Support Vector Machine

Support vector machine (SVM) [5] is a popular supervised learning method for classification and regression. SVM operates by first training a model (the training phase) followed by test/classification (the test phase). During the training phase, labeled feature vectors are used to train a model. During the test phase, SVM produces a predictive label when provided with a new (test) feature vector. SVM training can be formulated as the solution to the following optimization problem:

$$\min \tfrac{1}{2}\|\mathbf{w}\|^2 + C \sum_i \xi_i$$
$$s.t.$$
$$c_i(\mathbf{w}^T \mathbf{x}_i - \mathrm{b}) \geq 1 - \xi_i$$
$$\xi_i \geq 0$$

where $C$ is the cost factor, $\xi_i$ is the soft margin, $\mathbf{x}_i$ is the feature vector, $c_i$ is the label corresponding to the feature vector $\mathbf{x}_i$, $\mathbf{w}$ is the weight vector, and $b$ is the bias. It can be shown that the optimum weights $\mathbf{w}_o$ are represented as a linear combination of the feature vectors that lie on the margins, i.e., support vectors. Kernel trick [5] can be employed to realize non-linear decision boundaries.

### 2.3 Computational Error Model

Timing errors occur in data-path circuits whenever a storage element captures an incorrect logic value at its input. The probability of such an error event increases dramatically when circuits are operating in NTV [8], i.e., supply voltages $0.3\,\text{V} \leq V_{dd} \leq 0.7\,\text{V}$. The resulting timing errors are a complex function of the circuit state, inputs, architecture, and the process technology. A statistical model of such errors is therefore essential in order to understand the impact of errors on system performance.

In this paper, the computed outputs and timing errors are treated as random variables (RVs). We employ capital letters and small letters to denote a RV $Y$ and its realization $y$, respectively. We employ the following additive error model [19]:

$$\mathbf{y}_a = \mathbf{y}_o \oplus \boldsymbol{\eta} \tag{1}$$

where $\mathbf{y}_a = [y_{a,0}^b, ....y_{a,B-1}^b]^T$ is the $B$-bit observed (erroneous) output of a pipeline stage which is also a realization of the RV vector $\mathbf{Y_a} = [Y_{a,0}^b, ....Y_{a,B-1}^b]^T$, $\mathbf{y}_o = [y_{o,0}^b, ....y_{o,B-1}^b]^T$ is the ideal (error-free) output which is also a realization of the RV vector $\mathbf{Y_o} = [Y_{o,0}^b, ....Y_{o,B-1}^b]^T$, and $\boldsymbol{\eta} = [\eta_0^b, ....\eta_{B-1}^b]^T$ is the timing error vector which is also a realization of the RV vector $\mathbf{N} = [N_0^b, ....N_{B-1}^b]^T$, $\oplus$ is the element-wise addition in Galois Field over 2 (GF(2)), and $y_{a,i}^b, y_{o,i}^b, \eta_i^b \in \{0,1\}$ ($i = 0,\ldots,B-1$) are the $B$ bits of $\mathbf{y}_a, \mathbf{y}_o$, and $\boldsymbol{\eta}$, respectively.
During the error modeling phase, samples of the timing error $\boldsymbol{\eta}$ are obtained via HDL simulations. Using these samples, we estimate the parameters of the error probability mass function (PMF) $P(\boldsymbol{\eta})$ for use in system simulations. Since the logic errors are bit-level events that are made dependent by the structure of the logic network, the error bits follow a joint Bernoulli distribution. The procedure to obtain an analytical model for $P(\boldsymbol{\eta})$ begins by determining a latent Gaussian RV $\mathbf{U} = [U_0, ..., U_{B-1}]^T$ with mean vector $\boldsymbol{\mu}_u$ and covariance matrix $\mathbf{C}_u$ which are chosen such that $P(\boldsymbol{\eta})$ can be expressed as:

$$P(\boldsymbol{\eta}) = \Phi([0,...0]^T; \mathbf{D}\boldsymbol{\mu}_u, \mathbf{D}\mathbf{C}_u\mathbf{D}^T)$$

where

$$\mathbf{D} = \begin{pmatrix} (-1)^{\eta_0^b} & 0 & 0 \\ 0 & \ddots & 0 \\ 0 & 0 & (-1)^{\eta_{B-1}^b} \end{pmatrix}$$

and $\Phi(u; \boldsymbol{\mu}_u, \mathbf{C}_u)$ is the cumulative distribution function of a multivariate Gaussian with mean vector $\boldsymbol{\mu}_u$ and covariance matrix $\mathbf{C}_u$. The mean vector $\boldsymbol{\mu}_u$ and covariance matrix $\mathbf{C}_u$ are estimated from error samples obtained via HDL simulations.

During system simulations, we employ the following procedure to perform error injection: 1) generate samples $\mathbf{u} = [u_1, \ldots, u_{B-1}]^T$ of RV $U$, 2) employ the dichotomized Gaussian (DG) approximation [13] to generate $\eta_i^b$ from $u_i$ as follows:

$$\eta_i^b = \begin{cases} 1 & u_i \geq 0 \\ 0 & u_i < 0 \end{cases} \text{ for } (i = 0, 1..., B-1) \qquad (2)$$

and 3) perform error injection by employing (1). In this paper, the error model $P(\boldsymbol{\eta})$ and the error injection procedure above are employed in system simulations in Section 4.

## 3 System Architecture

In this section, we present system architectures for RF and SVM classifiers. The RF classifier is chosen as the implemented CE architecture due to its comparison based architecture which results in simple base learner architecture. The polynomial SVM is chosen as the implemented centralized architecture as it offers a good trade-off between decision boundary flexibility and hardware complexity [12].

### 3.1 The RF Architecture

The RF classifier is implemented using an ensemble of $L$ two-stage DT classifiers (weak learners) shown in Fig. 2(a). The $l^{th}$ DT is trained from a bootstrapped training set $\mathcal{S}_l$ obtained from the original training set $\mathcal{S}$, and processes the $M_l$-dimensional data vector $\mathbf{x}_l = [x_{l,1}, x_{l,2}, \ldots, x_{l,M_l}]^T$ obtained from the $M$-dimensional test data vector $\mathbf{x}$ ($M \gg M_l$).

Stage 1 of the $l^{th}$ DT consists of a comparator array that computes $sgn(x_{l,i} - T_{l,i})$ ($l = 1, 2, \ldots, L$, and $i = 1, 2, \ldots, M_l$) where $T_{l,i}$s are the thresholds obtained via training. Stage 2 consists of a look up table (LUT) which encodes the decision of each root-to-leaf path into a 1-bit output $y_{a,l} \in \{0, 1\}$. The outputs of the $L$ DTs are combined via a voter block to generate the final decision. Each DT is trained using the Gini index [3] as the training criterion.

Conventionally, a majority voter is employed to combine the outputs from all DTs as follows:

$$\hat{y}_a = maj(y_{a,1}, y_{a,2}, ..., y_{a,L})$$

where $\hat{y}_a \in \{0, 1\}$ is the majority voter output, and $y_{a,l}$ is the $l^{th}$ DT output given by:

$$y_{a,l} = y_{o,l} \oplus \eta_l$$

where $y_{o,l} \in \{0, 1\}$ is the error-free output and $\eta_l \in \{0, 1\}$ is the timing error of the $l^{th}$ DT. The RF with majority voter is denoted as RF-M. In case of binary classification, the majority voter can be implemented as shown in Fig. 2(b).

In order to enhance the robustness of RF in presence of timing errors, we propose an *error weighted voting* scheme where the timing error statistics are incorporated during the decision process. In order to do so, we employ the maximum-a-posterior (MAP) criterion, i.e.:

$$\hat{y}_a = \arg\max_{\forall c \in \mathcal{C}} P(c|\mathbf{x}) \qquad (3)$$

where $\mathcal{C}$ is the label set, $P(c|\mathbf{x})$ is the posterior probability of class label $c$ conditioned on the *test data* $\mathbf{x}$. Thus:

$$P(c|\mathbf{x}) = \sum_{l=1}^{L} P(c|R_l, \mathbf{x}) P(R_l|\mathbf{x}) \qquad (4)$$

$$= \sum_{l=1}^{L} P(c|R_l, \mathbf{x}) P(R_l) \qquad (5)$$

$$\approx \sum_{l=1}^{L} \mathbf{1}\{y_{a,l} = c\} p_l \qquad (6)$$

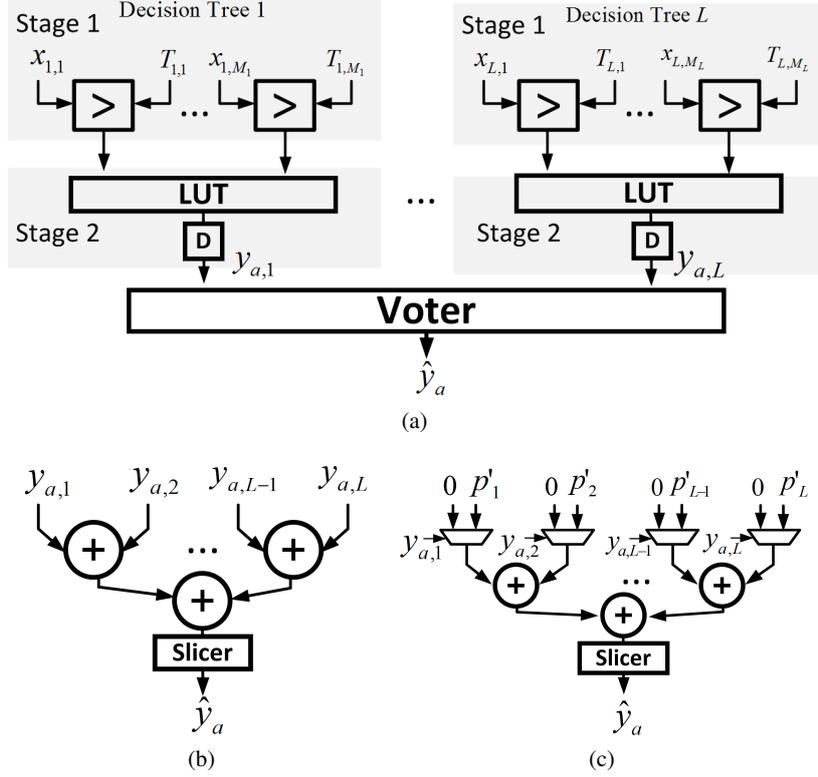

Figure 2: System architecture for: a) the RF classifier with $L$ DTs, b) the majority voter, and c) the weighted voter.

where $P(c|R_l, \mathbf{x})$ denotes the posterior probability of the class label, $R_l$ is the event of the $l^{th}$ DT being correct during the training phase, $p_l = P(R_l)$ is the probability of the event $R_l$, and $\mathbf{1}\{\cdot\}$ denotes the indicator function. Equation (4) implies (5) because the test data $\mathbf{x}$ and event $R_l$ are independent, and (5) implies (6) because we assume the DT output has a probability mass of 1 at the selected class label. The final decision $\hat{y}_a$ is obtained from (3) by choosing the label $c$ that maximizes (6). Note that $p_l$ represents the decision accuracy of the $l^{th}$ DT in presence of timing errors.

In the case of binary classification, one can simplify (3) using (6) into:

$$\hat{y}_a = \begin{cases} 1 & \text{if } \sum_{l=1}^{L} \mathbf{1}\{y_{a,l} = 1\} p'_l > \frac{1}{2} \\ 0 & \text{otherwise} \end{cases}$$

where $p'_l = \frac{p_l}{\sum_{l=1}^{L} p_l}$ and the voter can be implemented as shown in Fig. 2(c).

To incorporate the timing error statistics of each DT, we express $p_l$ in (6) as follows:

$$p_l = \sum_{\eta_l=0}^{1} P(R_l, \eta_l) = \sum_{\eta_l=0}^{1} P(R_l|\eta_l) P(\eta_l) \qquad (7)$$

where $P(R_l|\eta_l)$ is the probability of correct decision of the $l^{th}$ DT conditioned on $\eta_l$. The probabilities $P(R_l|\eta_l)$ and $P(\eta_l)$ can be obtained during the training phase for each DT. For a RF binary classifier, (7) can be simplified from the theorem of total probability as follows:

$$\begin{aligned} p_l &= P(R_l|\eta_l = 0) P(\eta_l = 0) + P(R_l|\eta_l = 1) P(\eta_l \neq 0) \\ &= P(R_l|\eta_l = 0)(1 - p_{\eta_l}) + P(R_l|\eta_l = 1) p_{\eta_l} \end{aligned} \qquad (8)$$

In binary classification, the erroneous output of the $l^{th}$ DT can be expressed as:

$$y_{a,l} = c \oplus \eta_l \oplus e_l \qquad (9)$$

where $c, \eta_l, e_l$ denote the true label, timing error, and error due to noise in data, respectively. Thus, the event $R_l = \{e_l \oplus \eta_l = 0\}$, and we have:

$$\begin{align}
P(R_l|\eta_l = 1) &= P(e_l \oplus \eta_l = 0|\eta_l = 1) \\
&= P(e_l = 1|\eta_l = 1) \tag{10} \\
&= P(e_l = 1) \tag{11} \\
&= P(e_l = 1|\eta_l = 0) \\
&= P(e_l \oplus \eta_l = 1|\eta_l = 0) \\
&= 1 - P(e_l \oplus \eta_l = 0|\eta_l = 0) \\
&= 1 - P(R_l|\eta_l = 0) \tag{12}
\end{align}$$

where (10) to (11) comes from the independence of $e_l$ and $\eta_l$. Substituting (12) into (8) leads to:

$$p_l = P(R_l|\eta_l = 0)(1 - p_{\eta_l}) + (1 - P(R_l|\eta_l = 0))p_{\eta_l} \tag{13}$$

where $P(R_l|\eta_l = 0)$ can be obtained via performing validation using out-of-bag samples, and $p_{\eta_l} = P(\eta_l \neq 0)$ is the error rate of the $l^{th}$ DT. As indicated in 13, the error weighted voting decreases the weight of the $l^{th}$ DT when its error rate increases. We denote RF with error weighted voting scheme as RF-EW.

When error rate $p_{\eta_l} = 0$, the error weighted voting scheme reduces to the conventional weighted voter [6] where $p_l = P(R_l|\eta_l = 0)$. The RF with conventional weighted voter is denoted as RF-W.

The performance of RF-EW improves when the DTs exhibit uncorrelated errors, i.e., the DT outputs exhibit diversity in terms of error statistics. It is possible to enhance DT diversity by designing each DT to have different: 1) algorithm (algorithmic diversity), 2) architecture (architectural diversity), and 3) data-path precision (precisional diversity), across the DT ensemble. Precision has a significant impact on the timing error statistics since the hardware errors under investigation are due to timing violations. Therefore, in this paper, the precision of each DT data-path in the RF-EW is randomly assigned uniformly between 4b and 8b, leading to different critical path delays among the DTs, and hence uncorrelated errors.

### 3.2 The SVM Architecture

The centralized machine learning algorithm employed in this paper is a second-order polynomial kernel SVM described as:

$$\begin{align}
\hat{y}_a &= sgn(y_a) \\
y_a &= \sum_{i=1}^{N} (\beta \mathbf{s}_i^T \mathbf{x} + \gamma)^2 \alpha_i + b \tag{14}
\end{align}$$

where $\mathbf{x} = [x_1, x_2, ..., x_M]^T$ is the $M$ dimensional test data vector, $\mathbf{s}_i = [s_1, s_2, ..., s_M]^T$ is the $i^{th}$ support vector, $\alpha_i$ is the weight associated with $\mathbf{s}_i$, $b$ is the bias, $\beta$ and $\gamma$ are parameters of the polynomial kernel, and $N$ is the total number of support vectors (typically $N \gg M$). Direct computation of (14) requires $O(NM)$ multiply-accumulate (MAC) operations. The following reformulation [12] reduces the number of MAC operations to $O(M^2)$:

$$\begin{align}
y_a &= \tilde{\mathbf{x}}^T \tilde{\mathbf{W}} \tilde{\mathbf{x}} + b \tag{15} \\
\tilde{\mathbf{W}} &= \sum_{i=1}^{N} \alpha_i \tilde{\mathbf{s}}_i \tilde{\mathbf{s}}_i^T
\end{align}$$

where $\tilde{\mathbf{W}}$ is a precomputed weight matrix, $\tilde{\mathbf{x}} = \begin{bmatrix} 1 \\ \mathbf{x} \end{bmatrix}$, and $\tilde{\mathbf{s}}_i = \begin{bmatrix} \gamma \\ \beta \mathbf{s}_i \end{bmatrix}$. Figure 3 shows a folded SVM architecture implementing (15) where Stage 1 computes $\tilde{\mathbf{W}} \tilde{\mathbf{x}}$, and Stage 2 computes the dot product between $\tilde{\mathbf{x}}$ and Stage 1 output, and adds the bias term $b$.

### 3.3 System Analysis

The potential robustness improvement achieved by RF can be analyzed by inspecting the generalized error $E\big[\big(C - \frac{1}{L}\sum_{l=1}^{L} \hat{Y}_{a,l}\big)^2\big]$ where $C$ is the label and $\frac{1}{L}\sum_{l=1}^{L} \hat{Y}_{a,l}$ is the RF output where equal weights in the voter is assumed for simplicity of analysis. Here the expectation is taken over the distribution of the label $C$, the training set $\mathcal{S}$, and the timing error $N_1, ..., N_L$.

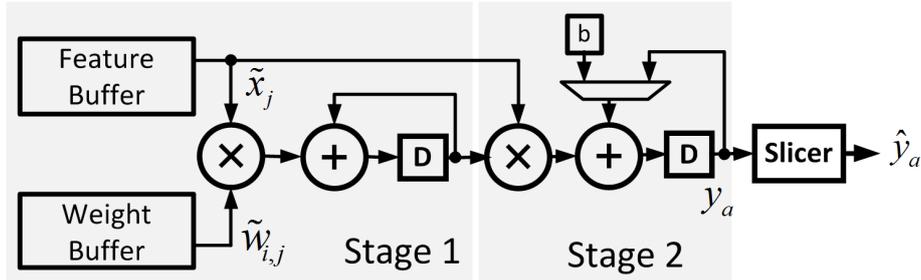

Figure 3: System architecture for a second-order polynomial kernel SVM classifier.

We start by deriving the generalized error for a single DT defined as $E[(C - \hat{Y}_a)^2]$ where $\hat{Y}_a$ is the DT output. It can be shown that (see Appendix A):

$$E\left[\left(C - \hat{Y}_a\right)^2\right] = \sigma_C^2 + b^2 + \sigma_{\hat{Y}_a}^2 \quad (16)$$

where $\sigma_C^2 = E\left[\left(C - E[C]\right)^2\right]$ is the irreducible error (noise), $b^2 = (E[C] - E[\hat{Y}_a])^2$ is the bias term and $\sigma_{\hat{Y}_a}^2 = E\left[\left(E[\hat{Y}_a] - \hat{Y}_a\right)^2\right]$ is the variance of $\hat{Y}_a$. Such a decomposition identifies the contribution of different error sources and allows one to understand the effect of CE in reducing these errors.

For CE, it can further be shown that the noise $\sigma_{C,RF}^2$ and the bias $b_{RF}^2$ (corresponding to the first two terms in (16)) do not change, i.e., $\sigma_{C,RF}^2 = \sigma_C^2$ and $b_{RF}^2 = b^2$, respectively. However, the output variance $\sigma_{RF}^2$ can be expressed as (see Appendix B):

$$\sigma_{RF}^2 = \frac{1}{L}\sigma_{\hat{Y}_a}^2 \quad (17)$$

We can see from (17) that $\sigma_{RF}^2$ is reduced by a factor of $\frac{1}{L}$ from $\sigma_{\hat{Y}_a}^2$, and that for the RF to achieve a lower variance, $\sigma_{\hat{Y}_a}^2$ should be less than $L$ times the variance of the centralized system. The reduction of variance leads to reduced generalized error and mis-classification rate.

## 4 Simulation Results

This section begins with the validation of the timing error model of Section 2.3 in a commercial $45\,\text{nm}$ CMOS process. These timing error models are derived for the RF and SVM classifiers of Fig. 2(a) and Fig. 3, respectively. Next, in Section 4.2, the detection accuracy of the SVM and RF architectures are compared using the validated error models. We employ the Breast Cancer Wisconsin dataset from UCI machine learning repository [1] which consists of labeled feature vectors (benign vs. malignant) constructed from digitized images of fine needle aspirates (FNA) of patient tissue. The SVM architecture being considered in this study consists of two types of MACs: Stage 1 employs $8\,\text{b}$ input, $8\,\text{b}$ coefficient, and Stage 2 employs $10\,\text{b}$ input, $8\,\text{b}$ coefficient MACs. The conventional RF-M and RF-W have Stage 1 consisting of comparator arrays with $8\,\text{b}$ input and $8\,\text{b}$ thresholds, and LUTs implemented as logic networks during the architecture generation. In the proposed RF-EW, each DT is implemented using a randomly selected precision uniformly distributed between $4\,\text{b}$ and $8\,\text{b}$ for both the input and the thresholds. These precisions were chosen to obtain less than $0.5\%$ degradation in $\hat{P}_{det}$ compared to a floating point implementation. The RF is trained by choosing randomly 3 features per node, and stopping the tree growth when the current node is pure or contains less than or equal to 2 samples. We do not restrict the depth of the tree. The complexities of the SVM and RF (with ensemble size $L = 10$) were found to be 1.63K and 1.47K 2-input NAND gate equivalents, respectively. In all cases, the parameters of SVM and RF were trained assuming no timing errors. During simulation, 50% of the data is employed during the training and testing, respectively.

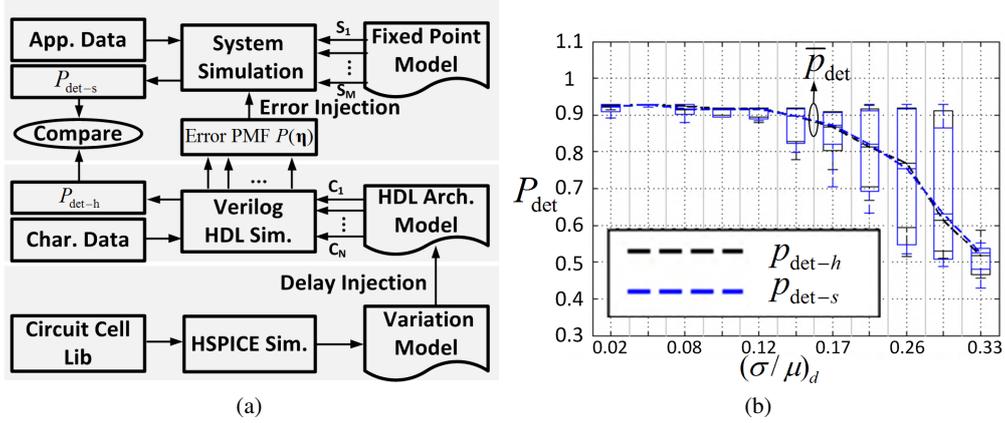

Figure 4: Error model generation and validation methodology: (a) model generation methodology, and (b) validation using the SVM architecture (averaged over 30 SVM instances) operating at gate level-delay variation of 2.8% - 33%.

### 4.1 Model Generation Methodology and Validation

The error model generation and validation methodology is shown in Fig. 4(a), and described below:

1. Characterize the gate delay distribution vs. operating voltage $V_{dd}$ of basic gates using HSPICE in NTV range $0.3\,\text{V}$-$0.7\,\text{V}$.
2. Implement the SVM and RF architectures shown in Fig. 3 and Fig. 2(a), respectively, using structural Verilog HDL using the basic gates characterized in Step 1.
3. Emulate process variations at NTV by generating multiple (30) architectural instances of each type (SVM and RF) and assigning random gate delays obtained via sampling the gate delay distributions obtained in Step 1. Note that the presence of process variation makes the detection accuracy $p_{det} = P(\hat{Y}_a = c)$ a RV, which we denote as $P_{det}$.
4. Run HDL (bit and clock accurate) simulations of each instance using a characterization dataset to obtain error samples $\boldsymbol{\eta}$ and classification accuracy $P_{det-h}$ for the two architectures. The characterization dataset is obtained via sampling with replacement from the application level data to emulate the input statistics.
5. Generate the error PMF $P(\boldsymbol{\eta})$ employing the procedure described in Section 2.3 [19].
6. Run fixed-point MATLAB simulations using $P(\boldsymbol{\eta})$ to inject errors for both SVM and RF using the UCI dataset to obtain detection accuracy $P_{det-s}$. Compare $P_{det-s}$ with $P_{det-h}$.

Figure 4(b) plots the SVM detection accuracy $P_{det-h}$ obtained in Step 4 (HDL simulations using gate delay distributions) and $P_{det-s}$ obtained in Step 6 (MATLAB simulations using $P(\boldsymbol{\eta})$) as a function of gate-level delay variation $(\sigma/\mu)_d$. We find that the median $P_{det-h}$ ($\bar{p}_{det-h}$) and $P_{det-s}$ ($\bar{p}_{det-s}$) differs by no more than 5% when $(\sigma/\mu)_d$ varies between 2.8% and 33%. Figure 4(b) also shows that the variation of $P_{det-h}$ increases as $(\sigma/\mu)_d$ increases from 2.8% to 29%, and then reduces because all the instances fail to perform correct classification for further increases in $(\sigma/\mu)_d$. The variation in $P_{det-h}$ is also modeled accurately as the maximum and minimum values of $P_{det-h}$ and $P_{det-s}$ differ by no more than 3% and 5%, respectively. Similar results were obtained for the RF architecture as well. These results indicate that the timing error is well-modeled by its PMF $P(\boldsymbol{\eta})$, and that the system performance can be accurately estimated by employing the methodology in Section 4.1.

### 4.2 Comparison of SVM and RF

#### 4.2.1 Comparison of timing error rates

We first compare the timing error rates $p_\eta = P(\boldsymbol{\eta} \neq 0)$ of SVM and RF obtained via HDL simulations as the voltage decreases in NTV. Figure 5(a) shows that the median timing error rate $\bar{p}_\eta$ increases by $500\times$ from $2.1 \times 10^{-3}$ to 0.99, and from $1.1 \times 10^{-3}$ to 0.61 for SVM and RF, respectively, as the voltage $V_{dd}$ decreases from $0.7\,\mathrm{V}$ to $0.3\,\mathrm{V}$, indicating that the RF architecture has up to $4.5\times$ lower timing error rate compared with SVM. The error rate of RF architecture is lower because it has comparator blocks which have a much simpler data path compared with the MAC units in SVM. Figure 5(a) also demonstrates that the gate level delay variation $(\sigma/\mu)_d$ increases by $12\times$ from 2.8% to 33% as the voltage $V_{dd}$ decreases from $0.7\,\mathrm{V}$ to $0.3\,\mathrm{V}$.

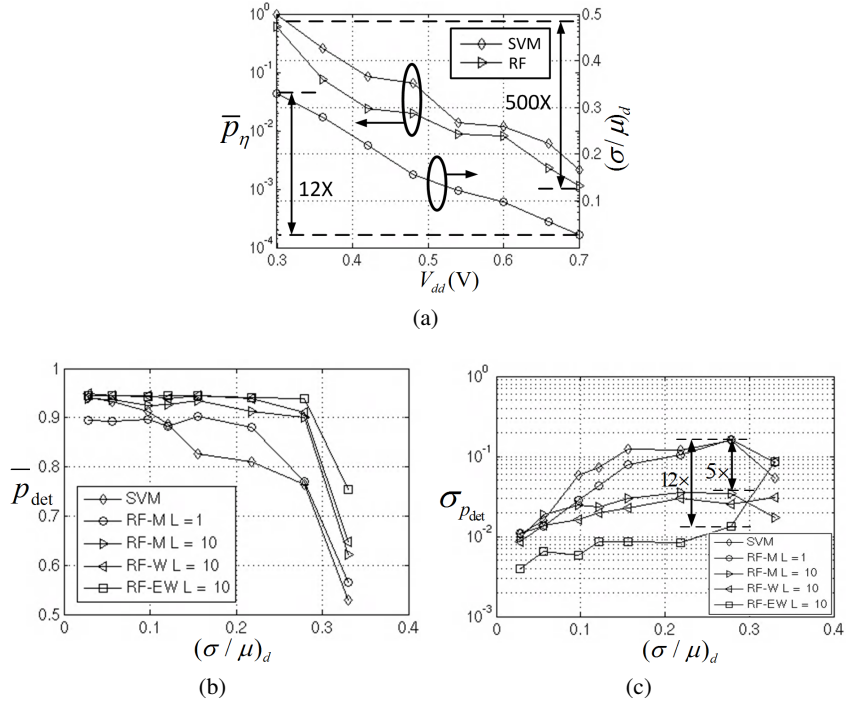

Figure 5: Robustness comparison in: a) error rate, b) the median detection accuracy $\bar{p}_{det}$, and c) the standard deviation of detection accuracy $\sigma p_{det}$ for SVM classifier, RF-M with $L = 1$ (i.e. signle DT), and RF-M ($L = 10$), RF-W ($L = 10$), and RF-EW ($L = 10$). Simulations were performed over 30 instances.

Next, we employ $P(\boldsymbol{\eta})$ to inject errors in fixed-point MATLAB simulations of SVM and RF architectures to compare their robustness to timing errors in NTV. All comparisons henceforth are in terms of $P_{det-s}$. Hence, we simplify the subscript and denote the detection accuracy as $P_{det}$. Four architectures are compared: 1) SVM, 2) RF with majority voter [3] (RF-M), 3) RF with weighted majority voter [6] (RF-W), and 4) RF with the proposed error weighted voter (RF-EW). We will compare the four architectures in terms of median ($\bar{p}_{det}$) and standard deviation ($\sigma_{p_{det}}$) of detection accuracy $P_{det}$.

### 4.2.2 Comparison of $\bar{p}_{det}$

Figure 5(b) shows that RF has higher $\bar{p}_{det}$ than SVM when the ensemble size $L$ is sufficiently large. Specifically, RF-M is able to maintain $\bar{p}_{det} \geq 0.9$ for $(\sigma/\mu)_d \leq 28.9\%$ with $L = 10$, whereas SVM can only maintain the same performance for $(\sigma/\mu)_d \leq 11.7\%$. Additionally, RF-EW achieves up to 3% higher $\bar{p}_{det}$ compared with RF-W and RF-EW, and is able to maintain $\bar{p}_{det} \geq 0.9$ for $(\sigma/\mu)_d \leq 29.6\%$. Finally, Figure 5(b) further shows that RF with $L = 10$ is able to maintain $\bar{p}_{det} \geq 0.9$ even at $(\sigma/\mu)_d$ of 28.9%. This indicates that RF architectures have a higher robustness to timing errors compared with SVM in spite of its complexity being lower by 10% when $L = 10$.

### 4.2.3 Comparison of $\sigma_{p_{det}}$

Figure 5(c) shows that $\sigma_{p_{det}}$ is significantly reduced as $L$ increases. RF-M achieves $\sigma_{p_{det}} \leq 3.5 \times 10^{-4}$ when $L = 10$, which is $5X$ lower compared to SVM or RF-M with $L = 1$. This further demonstrates that distributed architectures are inherently more robust to timing errors than centralized ones. Figure 5(c) also shows that RF-EW achieves $\sigma_{p_{det}} \leq 1.4 \times 10^{-2}$ when $(\sigma/\mu)_d \leq 29.6\%$, which is $12\times$ and $3.5\times$ lower compared to SVM and RF-W, respectively. This demonstrates that incorporating timing error statistics into the decision making process enhances robustness. When $(\sigma/\mu)_d \geq 30\%$, $\sigma_{p_{det}}$ of RF-EW is higher than that of RF-M and RF-W because all instances of RF-M and RF-W achieve a low $P_{det} \approx 0.6$, whereas some instances of RF-EW can still achieve a $P_{det} \geq 0.9$, leading to increased $\sigma_{p_{det}}$.

To understand the robustness improvement achi-eved by RF, Fig. 6 shows that the RF output variance $\sigma_{RF}^2$ reduces from 0.16 to 0.02 as $L$ increases from 1 to 25 when no precision diversity is employed. The variance reduction is more significant when the ensemble size $L$ is small, and slows down as $L$ further increases. This is because the independence assumption across the DTs is violated for large $L$. Figure 6 also shows that $\sigma_{RF}^2$ can be further reduced to 0.01 due to more uncorrelated error statistics when precision diversity is employed as in the RF-EW. As shown in (16), the reduction of variance leads to lower generalized error and higher $P_{det}$.

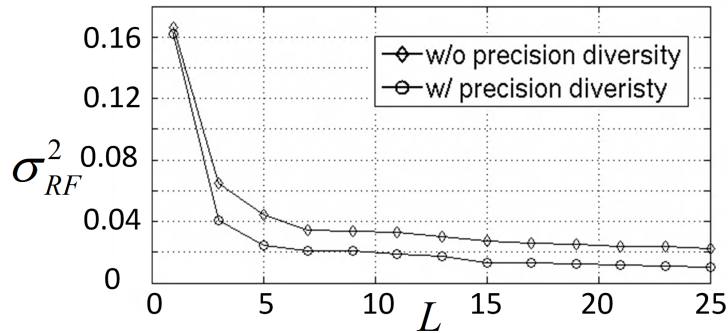

Figure 6: The variance of RF output when $(\sigma/\mu)_d = 29\%$.

## 5 Conclusion

In this paper, the inherent robustness of CE and centralized machine learning architectures in presence of timing violations is compared. It is shown that distributed architectures employing CE are inherently more robust than centralized ones to timing errors. Furthermore, it is shown that the algorithm itself can be adapted to further enhance the robustness. Such enhancement is achieved by using error weighted voting during the decision combination, and employing precision diversity in the architecture data path. The results demonstrate that in the CE framework, architectural level information can be incorporated at the system level to achieve enhanced robustness. In the future, architectural and algorithmic level diversity techniques can be employed to improve the robustness of CE. In addition, the robustness of CE in presence of defects errors (stuck-at-faults) can also be evaluated.

## Acknowledgment

This work was supported in part by Systems on Nanoscale Information fabriCs (SONIC), one of the six SRC STARnet Centers, sponsored by MARCO and DARPA.

## A

In this appendix, we derive (16). In deriving the generalized error $E[(C - \hat{Y}_a)^2]$, the expectation is taken over label $C$, the training set $\mathcal{S}$, and the timing error $N$. Here $\mathcal{S}$ and $N$ are independent. Without loss of generality, we assume a fixed input $X = \mathbf{x}$ as suggested by [7] for notational simplicity. Thus, (16) can be expressed as:

$$\begin{aligned} & E\big[(C - \hat{Y}_a)^2\big] \\ =\ & E\big[(C - E[C] + E[C] - \hat{Y}_a)^2\big] \\ =\ & E\big[(C - E[C])^2\big] + E\big[(E[C] - \hat{Y}_a)^2\big] \end{aligned} \qquad (18)$$

where we use the fact that

$$E\big[(C - E[C])(E[C] - \hat{Y}_a)\big]$$
$$= E\Big[E[(C - E[C])(E[C] - \hat{Y}_a)|S, N]\Big] = 0$$

The first term in (18) $E\big[(C - E[C])^2\big]$ is the noise $\sigma_C^2$. The second term in (18) can be further decomposed as:

$$E\big[(E[C] - \hat{Y}_a)^2\big]$$
$$= E\big[(E[C] - E[\hat{Y}_a] + E[\hat{Y}_a] - \hat{Y}_a)^2\big] \tag{19}$$
$$= (E[C] - E[\hat{Y}_a])^2 + E\big[(E[\hat{Y}_a] - \hat{Y}_a)^2\big] = b^2 + \sigma_{\hat{Y}_a}^2 \tag{20}$$

where in going from (19) to (20) we use the fact that

$$E\Big[(E[C] - E[\hat{Y}_a])(E[\hat{Y}_a] - \hat{Y}_a)\Big]$$
$$= (E[C] - E[\hat{Y}_a])E\big[E[\hat{Y}_a] - \hat{Y}_a]\big] = 0$$

This completes the proof of (16).

## B

In this appendix, we derive (17). In deriving the generalized error $E[(C - \frac{1}{L}\sum_{l=1}^{L}\hat{Y}_{a,l})^2]$, the expectation is taken over $C, S$, and the timing error of the $L$ DTs $N_1, \ldots, N_L$. The generalized error can be decomposed similar to (18) and (20) as follows:

$$E\big[(C - \frac{1}{L}\sum_{l=1}^{L}\hat{Y}_{a,l})^2\big] = \sigma_C^2 + b_{RF}^2 + \sigma_{RF}^2$$

where $\sigma_C^2 = E\big[(C - E[C])^2\big]$ is the noise term same as the first term in (18). Assuming $\hat{Y}_{a,l}$ are i.i.d with the same distribution as in the single DT $\hat{Y}_a$, the bias term $b_{RF}^2$ can be simplified into:

$$b_{RF}^2 = \big(E[C] - E[\frac{1}{L}\sum_{l=1}^{L}\hat{Y}_{a,l}]\big)^2 = \big(E[C] - E[\hat{Y}_a]\big)^2$$

which is the same as the first term in (20), and $\sigma_{RF}^2$ can be simplified as follows:

$$\sigma_{RF}^2 = E\Big[(E[\frac{1}{L}\sum_{l=1}^{L}\hat{Y}_{a,l}] - \frac{1}{L}\sum_{l=1}^{L}\hat{Y}_{a,l})^2\Big]$$
$$= \frac{1}{L^2}E\Big[\big(\sum_{l=1}^{L}(E[\hat{Y}_{a,l}] - \hat{Y}_{a,l})\big)^2\Big] \tag{21}$$
$$= \frac{1}{L^2}\sum_{l=1}^{L}E\Big[(E[\hat{Y}_{a,l}] - \hat{Y}_{a,l})^2\Big] = \frac{1}{L}\sigma_{\hat{Y}_a}^2 \tag{22}$$

where (21) to (22) comes from the assumption on the independence of $\hat{Y}_{a,l}$. This completes the proof of (17).